\RequirePackage{fix-cm}
\documentclass[smallextended]{svjour3}       
\smartqed  
\usepackage{graphicx}
\usepackage{algorithm}
\usepackage{algpseudocode}
\usepackage{pbox}
\usepackage{caption}
\usepackage{subcaption}
\usepackage{url} 
\usepackage{hyperref} 
\usepackage[section]{placeins} 
\usepackage{amssymb}
\usepackage{microtype}
\usepackage{hyphenat}

\begin{document}

\title{DeepStreamCE: A Streaming Approach to Concept Evolution Detection in Deep Neural Networks
\thanks{
\textbf{Code availability: http://github.com/chambai/deepstreamce}\\
}
}

\titlerunning{DeepStreamCE}        

\author{Lorraine Chambers         \and
        Mohamed Medhat Gaber \and
        Zahraa S. Abdallah 
}


\institute{L. Chambers \at
              Birmingham City University \\
              \email{lorraine.chambers@mail.bcu.ac.uk}           
           \and
           M. Gaber \at
              Birmingham City University \\
              \email{mohamed.gaber@bcu.ac.uk} 
           \and
           Z. Abdallah \at
              Birmingham City University \\
              \email{zahraa.abdallah@bcu.ac.uk}
}

\date{Received: date / Accepted: date}

\maketitle

\begin{abstract}
Deep neural networks have experimentally demonstrated superior performance over other machine learning approaches in decision-making predictions. However, one major concern is the closed set nature of the classification decision on the trained classes, which can have serious consequences in safety critical systems.  When the deep neural network is in a streaming environment, fast interpretation  of this classification is required to determine if the classification result is trusted. Un-trusted classifications can occur when the input data to the deep neural network changes over time. One type of change that can occur is \emph{concept evolution}, where a new class is introduced that the deep neural network was not trained on. In the majority of deep neural network architectures, the network only has the option to assign this instance to one of the classes it was trained on, which would be incorrect. The aim of this research is to detect the arrival of a new concept/class in the stream and indicate that to the user of the system. Existing work on interpreting deep neural networks  often focuses on visual interpretation and feature extraction with regards to which neurons have activated for certain input features. In this research, we propose fast deep neural network interpretation for streaming data to detect concept evolution. Our novel approach, coined \emph{DeepStreamCE}, uses streaming approaches for real-time Concept Evolution detection in deep neural networks. \emph{DeepStreamCE} applies neuron activation reduction using an autoencoder and MCOD stream-based clustering in the offline phase. Both outputs are used in the online phase to analyse the neuron activations in the evolving stream in order to detect concept evolution occurrence in real time. We evaluate \emph{DeepStreamCE} by training VGG16 convolutional neural networks on various combinations of data from the CIFAR-10 dataset, holding out some classes to be used as concept evolution.  For comparison, we apply the data and VGG16 networks to an open-set deep network solution -- OpenMax. \emph{DeepStreamCE} outperforms OpenMax when identifying concept evolution for our datasets.
\end{abstract}

\section{Introduction}
\label{sec:intro}
The ability of deep neural networks to classify data based on a sufficiently representative training set is established experimentally.  However, when the unseen instances are presented and they deviate from the training set distribution, they could be incorrectly classified.  This is problematic in safety critical systems such as autonomous vehicles, flight control, medical image classification or medical sensor analysis.  In such systems, data would be arriving real-time and any potentially incorrect classifications need to be captured as quickly as possible.  There are many reasons that the unseen instances can vary from the training data, ranging from valid data changes over time to deliberate adversarial attacks. The data discrepancy that this research focuses on is concept evolution, where new valid classes appear over time in the data stream.  

Detecting concept evolution in data streams in not a new topic. Various approaches have been discussed in the literature to detect evolution  in the stream ~\cite{masud_classification_2011,haque_efficient_2016,haque_sand_2016}. These approaches focus mostly on inspecting changes in the input data distribution using methods such as statistics~\cite{song_statistical_2007} or PCA~\cite{kuncheva_pca_2012}, or applying the data to one or more classifiers and look for a change in confidence of those classifiers. Since deep neural networks frequently have high dimensional input data,  detecting distributional change in such a multi-dimensional space is challenging~\cite{harel_concept_2014}. Hence, the latter option of using classifier outputs to detect change is more viable. However, classifier outputs do not leverage the feature rich activation data available within the deep neural networks. Therefore, in our research, we are detecting concept evolution specifically within data provided to deep neural networks which means we have the opportunity to change the input data space by utilising activations from the hidden layers within the deep neural network instead of the input data.  This research utilises a deep neural network that classifies images, known as a Convolutional Neural Network (CNN).  The network is given an image and it will calculate transformations on the image until it produces a classification label of the image.  Inside the network, there are hidden layers containing neurons which are functions that have weights and biases whose values are calculated during training of the network.  When the trained network processes the image that is given to it, values are produced in the hidden layers of the network.  These values are called activations.  The activations can be accessed and provide different information about the image.  As each hidden layer of the network has learned to detect different features of the image, this means that when we are looking at the activations, we are looking at the feature space of the image instead of the pixel space. This is beneficial as it makes the analysis independent of the type of input data and increases the space between closed set and open set instances~\cite{bendale_towards_2016}.

Analysis of the internal neuron activation of deep neural network is a popular area of study in the field of deep neural network inspection, visualisation and explainable AI \cite{buhrmester_analysis_2019,adadi_peeking_2018,carter_activation_2019,kahng_activis:_2018} and shows that activations of a deep neural network can be used to help determine how a network arrives at its classification decision.  Many deep neural networks operate within a streaming environment however, to our knowledge, there have been no studies analysing the activation data with streaming analysis techniques. 

Detecting and analysing neural network activations is challenging in a streaming environment due to the amount of activations, even for a low-resolution image and a small deep neural network.  Identification of the most important neurons based on their activation data is therefore required. We discuss an overview of existing techniques for detecting new classes in deep neural networks, existing techniques for extracting important neuron activation data and concept evolution detection in a streaming environment.  We propose utilising multi-layer activations from the deep neural network, reducing the activations via an autoencoder and Micro-cluster-based Continuous Outlier Detection (MCOD) \cite{kontaki_efficient_2016} based on streaming clustering for activation analysis to determine if concept evolution has occurred.
	
Our contributions can be stated as follows.
\begin{itemize}
    \item We detect the activation difference between unseen instances of concept evolution with the training data using streaming techniques;
    \item We use fast interpretation of deep neural network activations to detect concept evolution using streaming techniques;
    \item  We investigate how changing the streaming analysis parameters affects the concept evolution detection; and
    \item we compare our technique to a leading deep neural network open-set classification solution -- OpenMax \cite{bendale_towards_2016}.
\end{itemize}

This paper is organised as follows: In Section~\ref{sec:relwk} we first discuss the related work, then present a system description including formalisation and implementation details of  of the \emph{DeepStreamCE} components and methodology in Section~\ref{sec:desc}.  In the experimental study in Section~\ref{sec:exp}, we evaluate and analyse \emph{DeepStreamCE} on sub-datasets from the CIFAR-10 dataset and experiment with varying the input parameters of the streaming clustering algorithm.  The same data and deep neural networks are applied to an open-set deep network solution and the results are compared. Section~\ref{sec:concl} summarises our findings with a conclusion and future work suggestions.

\section{Background and Related Work}
\label{sec:relwk}
Within the field of deep neural networks, there are two areas of research that focus on how new classes are identified that the deep neural network has not been trained on: (1) open-set classification and (2) out-of-distribution detection.  
Open-set classification means that the deep neural networks have the ability to reject the unseen instances as unknown, rather than having to choose a classification from the known classes they were trained on \cite{bendale_towards_2016}.
Open-set classification deals with Rubbish/Fooling images -- images that are plainly rubbish to the human eye (i.e. computer generated patterns to adversarial images that are deliberately slightly modified images).  These represent the two opposite ends of the scale.  The rubbish/fooling images are far away from the feature space of the images the network was trained on and are easier to capture and  adversarial images that only manipulate a few pixels are close to the feature space of the original training instances.  Common methods for detecting images that differ from the training data are thresholding softmax scores, uncertainty estimation and extra training using negative samples as summarised by \cite{dhamija_reducing_2018}.  Other methods include using Extreme Value Theory (EVT) which is a branch of statistics dealing with extreme deviations from the median of probability distributions \cite{geng_recent_2019} and re-training the neural network using a different error loss function \cite{hassen_learning_2018}. Bendale proposes a system called OpenMax where activation patterns in the penultimate activation layer are utilised, and another probability calculation layer is added that then compares outputs with the original softmax layer of the network. The author also suggests that there could be a layer in the deep neural network where the activations will be far away from the training samples, where unknown images become outliers in an open set recognition problem \cite{bendale_towards_2016}.   The solutions used in out-of distribution involve perturbing the images and using thresholds and temperature scaling on the Softmax layer, or training the data.  Temperature scaling and adding small peterbations to images are used in \cite{liang_enhancing_2018} and in \cite{devries_learning_2018}, the penultimate layer output is used to calculate a confidence estimate for each data input into the deep neural network.  In the open-set classification and out-of-distribution fields, a common theme is using the activations of the layer preceding the Softmax layer. As the deep neural network activations represent the features of the image, it could provide more information than just using the penultimate layer activations or the pixel data.  Another field that uses activations is Deep Neural Network (DNN) Inspection, which is reviewed in Section~\ref{sec:relwk_acts}.

\subsection{Neuron Activations}
\label{sec:relwk_acts}
Activations have been widely used in the Visual Interpretation of deep neural networks to determine what neurons are related to what image features to explain how the neural network is arriving at its classification.  How the network arrives at its classification is out of the scope of this research as we are only interested in identifying the important neurons in an image’s classification.  Activations have also been used in the field of adversarial attacks on deep neural networks.  We draw inspiration from these fields with respect to the detection of important neurons.  

In the DNN inspection field, there have been many approaches to the identification of the most important activations and these have been recently surveyed, showing that this is an important area and forms part of explainable AI \cite{adadi_peeking_2018}.  In \cite{buhrmester_analysis_2019}, Table 1 shows that neuron activations are used for explainers for deep neural networks in different fields such as images and text based DNNs.  Approaches to using neuron activations can be summarised as: top $k$ percent of activations in each layer, the activation magnitude, average activations and clustering, nearest neighbour and backpropagation.  The first approach of top $k$ percent activations is used in \cite{hohman_summit:_2019}, which uses the activation of channels in a CNN, to determine edges, shapes and texture, and applies global max pooling to reduce the data.  It uses the activations of the channels and is appropriate to CNNs only.  This method only does a forward pass through the network to obtain the activations, so is low on computation, which is required in a streaming environment.  Activation Magnitude and Matrix Factorisation are utilised by Olah \cite{olah_building_2018} and uses the magnitude of the neuron activations and represents them as a cube and breaks them up using matrix factorisation to get more meaningful groups of neurons, however matrix factorisation is computationally expensive as it has to be done differently for each image, so is not suitable for a streaming environment.  Average activation and clustering is used in \cite{liu_towards_2017}, where the average activation of each neuron in the activation layer is used (average is taken for all instances with the same class), then clustered and a number of neurons from each cluster is selected.  In ActiVis \cite{kahng_activis:_2018}, the average activation for each neuron for all instances in a class are used, but presented to the user for visualisation.  The nearest neighbour approach was used in \cite{papernot_deep_2018} where nearest neighbour was used on the activation outputs of each hidden layer. Locality Sensitive Hashing (LSH) function is used to reduce the data dimensionality, so it is suitable for use in the nearest neighbour representation. However, this is computationally expensive and unsuitable for a streaming environment.  Backpropagation is used in \cite{samek_evaluating_2017}, \cite{qiu_adversarial_2019}. The latter is applicable to both CNNs and fully connected networks. This describes an effective critical path of weights and neurons that lead to the final predicted path and uses an activation-based back propagation algorithm to extract the effective path.  This requires a backward pass through the network which is computationally expensive and not appropriate in a streaming environment.

Activation data is also used in the adversarial detection field and although this research is concerned with detecting concept evolution, it is worth noting the work using activations in the adversarial detection field.  The work in \cite{chen_detecting_2018} clusters the activations of the last hidden neural network layer, flattened into a 1D vector and clustered. Dimensionality reduction is performed using Independent Component Analysis (ICA) to avoid issues with clustering on very high dimensional data -- as dimensionality increases, distance metrics are less effective.  They used k-means with $k=2$.  They train a a new model on the original data minus the data corresponding to the clusters in question, they use this model to classify the removed clusters.  If a cluster contained legitimate data then the cluster will be classified as its correct label, this is computationally expensive. In \cite{chen_detecting_2018}, the proposed method also uses neuron activation via back propagation, however, backpropagation is also computationally expensive. Hendrycks uses abnormality detection, using activations and suggests using auxiliary decoders such as autoencoders as further work \cite{hendrycks_baseline_2018}.

The usage of neuron activations faces the challenge of deciding which neurons are used (i.e. only use a particular layer(s) or channel(s), or general reduction of the activation such as in LSH to be suitable for post processing). Using the last activation layer only is commonly used as a method of data reduction as this is the most representative of the image and provides the most information.  However, we are not restricted to only using the last layer, and given recommendations on using activations with auxiliary decoders, a summary of data reduction techniques follows.

\subsection{Data Reduction}
\label{sec:relwk_datred}
Popular methods of data reduction are Independent Component Analysis (ICA), Principle Component Analysis (PCA), autoencoders and Restricted Boltzman Machines (RBMs).  The proposed method in \cite{chen_detecting_2018} uses ICA, however, Hinton describes using autoencoders as better than PCA \cite{hinton_reducing_2006}.  PCA and ICA are for linear transformations but when we are looking at DNNs, they are not linear transformations. For this research, autoencoders will be used as they provide scope for more complex data reduction, including expansion into RBMs.  Once the data is reduced we have the opportunity to use more detection methods on it.  As the aim is to detect concept evolution, we first review Concept Evolution detection techniques.

\subsection{Concept Evolution in Data Streams}
\label{sec:relwk_ce}
Concept evolution is the appearance of new classes while streams evolve. New concepts need to be detected as soon as they arrive, without being trained with labelled data. There has been much investigation into concept evolution \cite{gamajoao_survey_2014,faria_novelty_2016,khamassi_discussion_2018,masud_classification_2011,haque_efficient_2016,haque_sand_2016,abdallah_anynovel:_2016}, some of which cover concept drift, of which concept evolution is defined as one of its manifestations.  They also cover the whole process of concept evolution in data streams including the forgetting of the concept evolution and the update of the algorithm.  The aspects that are of interest to this research is the learning process that is used.  

There are two types of methods for detecting drift -- sequential methods and Windowing.  
Sequential methods are concerned with only one instance at a time.  It utilises statistical analysis and checks if there is a change between distributions of instances. One method is to measure the dissimilarity between an incoming instance and a set of data.  To measure this, distance function measures can be used  \cite{tran_distance-based_2016,tsymbal_problem_2004,goncalves_comparative_2014}, or summarised statistics from the two distributions like mean and variance \cite{ross_exponentially_2012}. In sequential approaches, each instance is processed only a single time, then discarded.  This is suitable for detecting drifts where data streams are infinite and it is not practical to store all the instances, which is usually the case in real-world applications.  These sequential approaches do detect abrupt drift, which is what a new class would be.  Windowing methods consider that the most recent observations are the most informative.  They progressively  estimate the change through a time or data window. Generally the windowing approach considers that the drift is uniform and affects the entire instance space, so they can handle global concept drifts -- this is sufficient for concept evolution. For this research, an unsupervised
windowing method has been selected to identify concept evolution via outlier detection. MCOD \cite{kontaki_continuous_2011} is an established outlier detection technique which performs clustering on continuous data streams \cite{haidar_data_2019} and it outperforms other streaming clustering methods \cite{tran_distance-based_2016}. Section~\ref{sec:desc_analy} provides a description of MCOD.  

In summary, deep neural networks produce many activations that require reducing before they are suitable to be used in outlier detection techniques to detect concept evolution. Section \ref{sec:desc} details our proposed methods for activation reduction and streaming outlier detection.  We use a multi-layer technique to extract activation information from the deep neural network. As this provides a large amount of data, an autoencoder approach is to be used to reduce this data. This will then be fed into a streaming clustering algorithm to detect outliers.

\subsection{Comparison with OpenMax}
\label{sec:relwk_omax}
This research will compare results with Bendale's OpenMax solution for detecting unknown classes \cite{bendale_towards_2016}.  OpenMax was selected for comparison as, similarly to \emph{DeepStreamCE}, it utilises activation data from within the deep neural network and identifies unknown instances. It uses the penultimate and final layer activations from the network. The final layer of the network are the Softmax probabilities and the penultimate layer is a representation of the instance's class. OpenMax estimates the probability of an instance being from an unknown class. To achieve this, it extracts the activations from the penultimate layer of the network, calculates the mean of these activations for each instance in the training data and constructs a vector of these for each class. For each training instance, it also calculates the distance between the instance and its class activation vector. For an unseen instance, the mean activation of the penultimate layer is calculated, the distance between the instance's mean activations and each class mean activations is measured, then a Weibull fit is applied to the distances between the instance and the classes and extreme value theory (EVT) is used to estimate the probability of the instance being an outlier with respect to each class. Either a known class or an `unknown' classification is returned. 

\section{DeepStreamCE System Description}
\label{sec:desc}
\begin{figure}[!htb]
\centerline{\includegraphics[width=\textwidth]{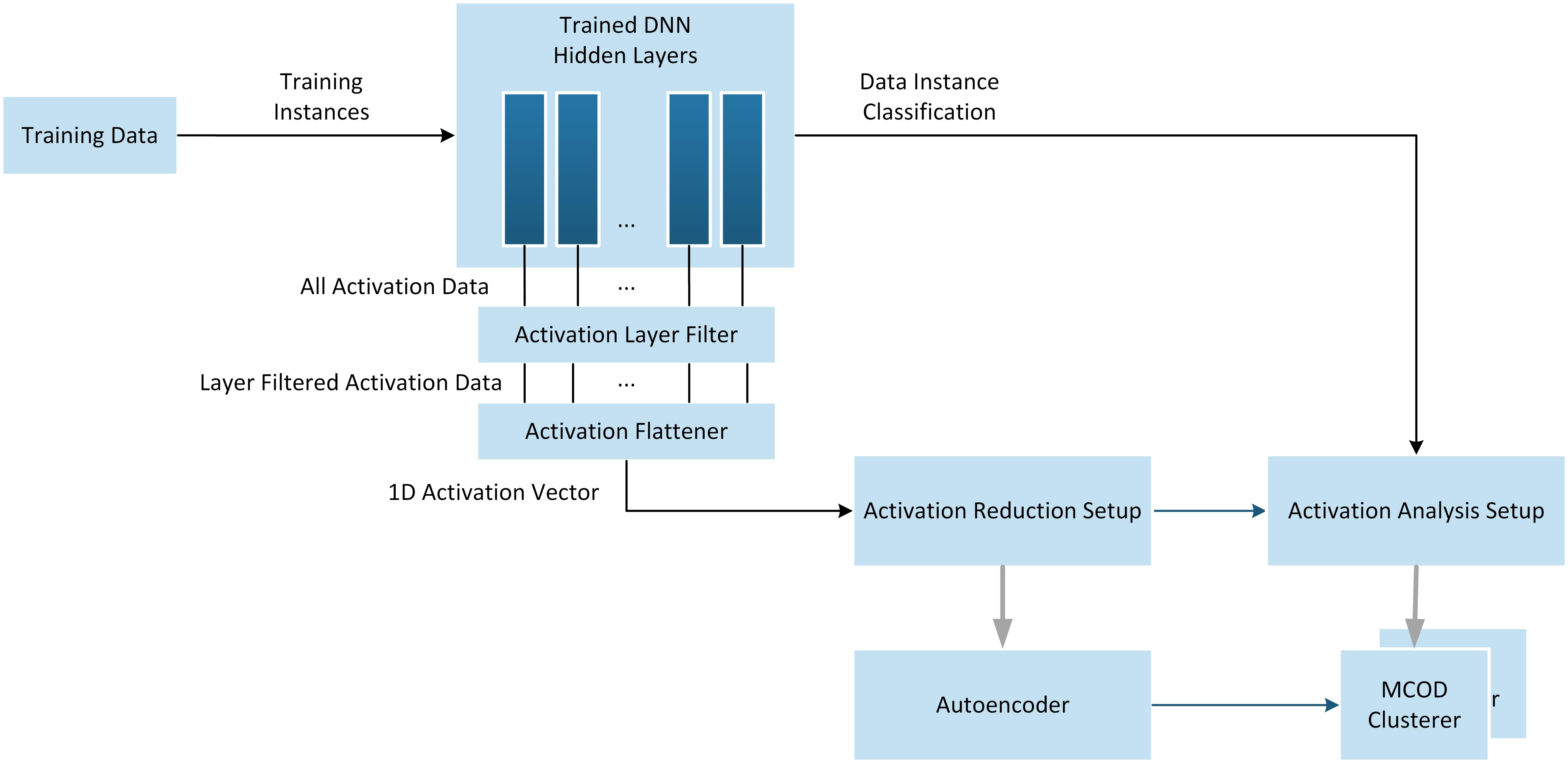}}
\caption{DeepStreamCE: Offline Phase}
\label{fig1}
\end{figure}

\begin{figure}[!htb]
\centerline{\includegraphics[width=\textwidth]{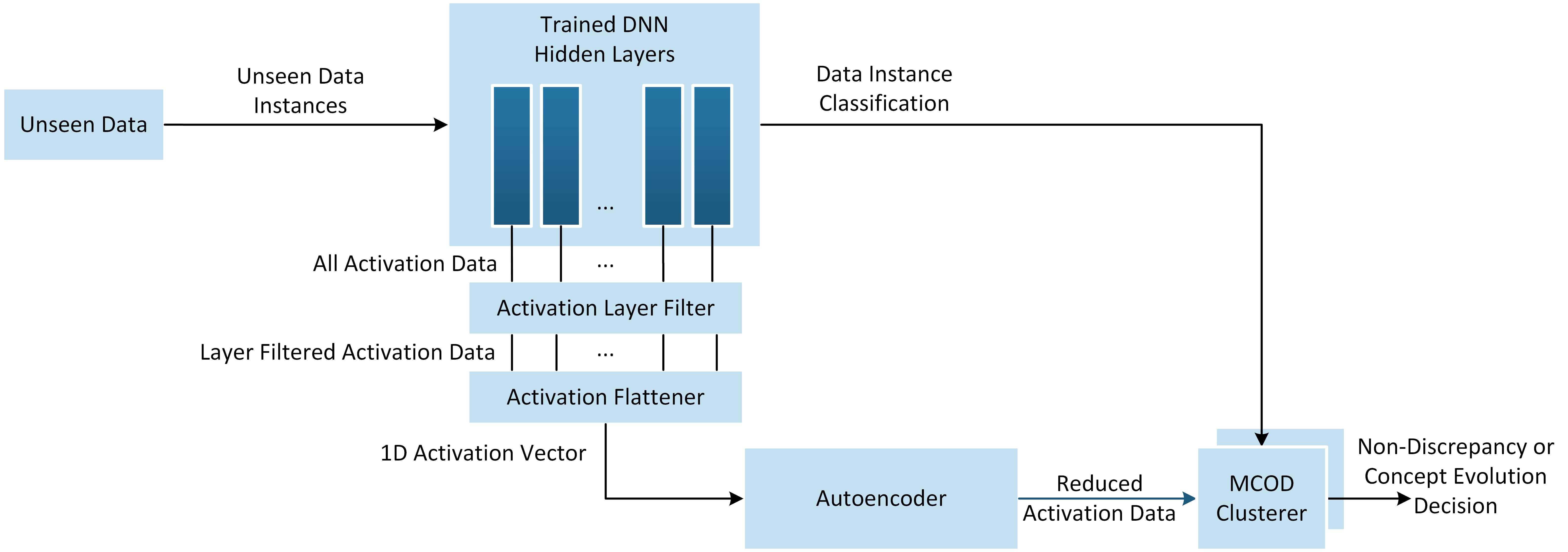}}
\caption{DeepStreamCE: Online Phase}
\label{fig2}
\end{figure}

DeepStreamCE is comprised of two stages, the offline phase and the online phase.  Figure~\ref{fig1} shows the offline phase components and Figure~\ref{fig2} shows the online phase components.  Prerequisites for the system are: (1) a trained deep neural network that is being analysed for concept evolution and (2) the data instances that the network was trained on.

\subsection{Offline Phase}
\label{sec:desc_offPh}
During the offline phase, all of the training instances are presented to the deep neural network and the resulting activations of each training instance are extracted. After this, the activation data goes through 3 stages: activation layer filtering, activation reduction setup and activation analysis setup. The algorithm for the offline phase is listed in Algorithm~\ref{algOff} and Table~\ref{tab2} shows the description for the \emph{DeepStreamCE} symbols.

\subsection{Activation Layer Filtering Setup}
\label{sec:desc_filt}
One data instance produces many activations. The amount of activations per data instance depends on the neural network architecture (e.g. how many layers and what kind of layers). For instance, if the network has activation layers  or fully connected layers as opposed to convolutional layers, there will be more activation data. Therefore, filtering of the activations needs to occur in order to proceed with a manageable amount of data, with the manageable size depending on the amount of system memory available. There is also a decision to be made regarding which layers are selected from the deep neural network. It is generally considered that the latter layers of a deep neural network produce more interesting data as it is closer to the final class outcome. For the network in this experimental setup, more information regarding the layer selection is given in Section~\ref{sec:exp_dnn}. The amount of memory required in the offline phase is important, and is where the maximum amount of memory is required to handle the activations extracted from the training data. The activations no longer represent the pixel space, but a representation of the pixel space, so we no longer need to keep the dimensions of the data, thus we flatten the activation data into a 1D vector.  (\textbf{flatten} -- line 2 of Algorithm~\ref{algOff}), ready for use in the activation reduction setup phase, as described in section ~\ref{sec:desc_red}.

\subsection{Activation Reduction Setup}
\label{sec:desc_red}
The activation data is high dimensional.  To be able to make use of this in a clustering algorithm requires that the dimensionality is reduced.  To do this, \emph{DeepStreamCE} uses an autoencoder to reduce the data to 100 dimensions. The autoencoder is trained using the data as selected from the activation layers of the deep neural network (line 4 of Algorithm~\ref{algOff}), then each training instance is processed through the autoencoder to reduce its dimensionality to 100 (\textbf{reduce} - line 6 of Algorithm~\ref{algOff}). The training and creation of the autoencoder has the largest memory and computational requirement of the system. For these initial experiments, the autoencoder is an undercomplete autoencoder~\cite{rumelhart_learning_1986} with a relu activation function and a mean squared error loss function, which makes it equivalent to PCA, This has potential for further work~\cite{hinton_reducing_2006,wang_auto-encoder_2016}. With this experimental setup, the autoencoder reduces the activations from 47104 to 100 dimensions. The activation instances are now ready for use in the activation analysis setup. 

\subsection{Activation Analysis Setup}
\label{sec:desc_analy}
\begin{figure}[H]
\centerline{\includegraphics[width=0.5\textwidth]{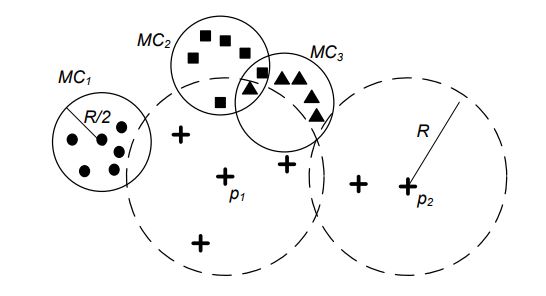}}
\caption{Example MCOD clusters for k=4}
\label{fig3}
\end{figure}

Stream clustering underpinning MCOD is used for the activation analysis \cite{kontaki_continuous_2011}.  Figure~\ref{fig3} shows the concept of MCOD and Table~\ref{tab1} describes the symbols used.  MCOD is based on a micro-clustering technique that takes parameters of: (1) radius (${R}$) of the micro-cluster (${MC}$), (2) the minimum number of instances to form a micro-cluster (${k}$) and (3) window size -- the number of instances considered in the clustering algorithm (${W}$). 

\begin{table}[H]
\centering
\caption{MCOD Symbols}
\begin{tabular}{ll}
\hline\noalign{\smallskip}
\label{tab1}
Symbol & Description\\
\hline\noalign{\smallskip}
$k$&Minimum number of neighbours to form a micro cluster\\
$R$&Radius - distance parameter for outlier detection\\
$MC$&Micro cluster\\
$M$&MCOD clusterer\\
$p$&data instance\\
$W$&Window size\\
\hline\noalign{\smallskip}
\end{tabular}
\end{table}

If there are $k + 1$ instances of $p$ within $R/2$ of an $MC$, $p$ becomes a member of that $MC$.  If $p$ is within $2R/2$ of any cluster, it becomes an outlier of those $MCs$. MCOD uses the centre of the micro clusters to perform its calculations, which makes it computationally efficient.

For the \emph{DeepStreamCE} implementation, one MCOD clusterer ($M$) is created for each possible classification output of the trained deep neural network (\textbf{createClusterer} -- line 9 of Algorithm~\ref{algOff}) and the reduced activation instances are added to the MCOD clusterer that represents their classification label (\textbf{addToClusterer} -- line 12 of Algorithm~\ref{algOff}). When the reduced instances are being added to the MCOD clusterer, micro clusters ($MC$) may be formed within the MCOD clusterer ($M$), however, during the offline phase we are not interested in these micro clusters as we do not require inlier/outlier decisions.  $W$ is set to the number of training instances for that class plus one and the effect of varying $R$ and $k$ is investigated in this research. These parameters can only be set once, at the time the MCOD clusterer is created. The use of the micro clusters for outlier detection in \emph{DeepStreamCE} is described in Section~\ref{sec:desc_onph}.

The output components from the offline phase is an autoencoder ($A$) and an MCOD clusterer for each class of the deep neural network ($M_j$). These are both used during the online phase, as described in Section~\ref{sec:desc_onph}.

\begin{table}[H]
\centering
\caption{DeepStreamCE Symbols}
\begin{tabular}{ll}
\hline\noalign{\smallskip}
\label{tab2}
Symbol & Description\\
\hline\noalign{\smallskip}
$I$&Number of instances\\
$i$&Instance iterator\\
$n$&Number of activation layers\\
$N$&Number of activation values\\
$v$&Activation value\\
$l$&Activation layer\\
$x$&Correctly classified training instance\\
$S$&Activation values for all layers in an instance\\
$f$&Flattened activations for an instance\\
$r$&Reduced activations for an instance\\
$j$&Non-discrepancy class\\
$A$&Trained autoencoder\\
$u$&Unseen instance\\
\hline\noalign{\smallskip}
\end{tabular}
\end{table}

\begin{algorithm}
	\caption{Offline Algorithm}
	\label{algOff}
\begin{algorithmic}[1]
        \Require {Pre-trained VGG16 DNN on 2 classes}
        \Require {Activation Levels in layers 9,12,13,15,16,17,20,21 layer} $\mathbf{v({x}_{i})} = {v}_{l1}(x)_{i}...{v}_{lN}({x}_{i})$
        \Require {MCOD Parameters ${k}, {R}$}
        \Require {For each correctly classified training example, $\mathbf{x}_{i}$ let $\mathbf{S}_{i}={v}_{l1}(x)_{i}...{v}_{lN}({x}_{i})$}
		\For {${l}$ = 1\dots${N}$}
		\State ${f}_{i}$ = \textbf{flatten}(${S}_{i}$)
		\EndFor
		\State Train autoencoder, ${A}$
		\For {${x}$ = 1\dots${I}$}
		\State Reduce activations to 100: ${r}_{i}$ = \textbf{reduce}(${A}$,${f}_{i}$)
		\EndFor
		\For {${j}$ = 1\dots${J}$}
		        \State Create Stream and MCOD clusterer: ${M}_{j}$ = \textbf{createClusterer}(${r}_{ij}$)
		\EndFor
		\For {${r}$ = 1\dots${I}$}
		    \State Add class instances to stream: ${M}_{j}$ = \textbf{addtoClusterer}(${M}_{j}$,${r}_{ij}$)
		\EndFor
		\State\Return{${A}$,${M}_{j}$}
	\end{algorithmic} 
\end{algorithm}

\subsection{Online Phase}
\label{sec:desc_onph}
\begin{figure}[H]
\centerline{\includegraphics[width=0.75\textwidth]{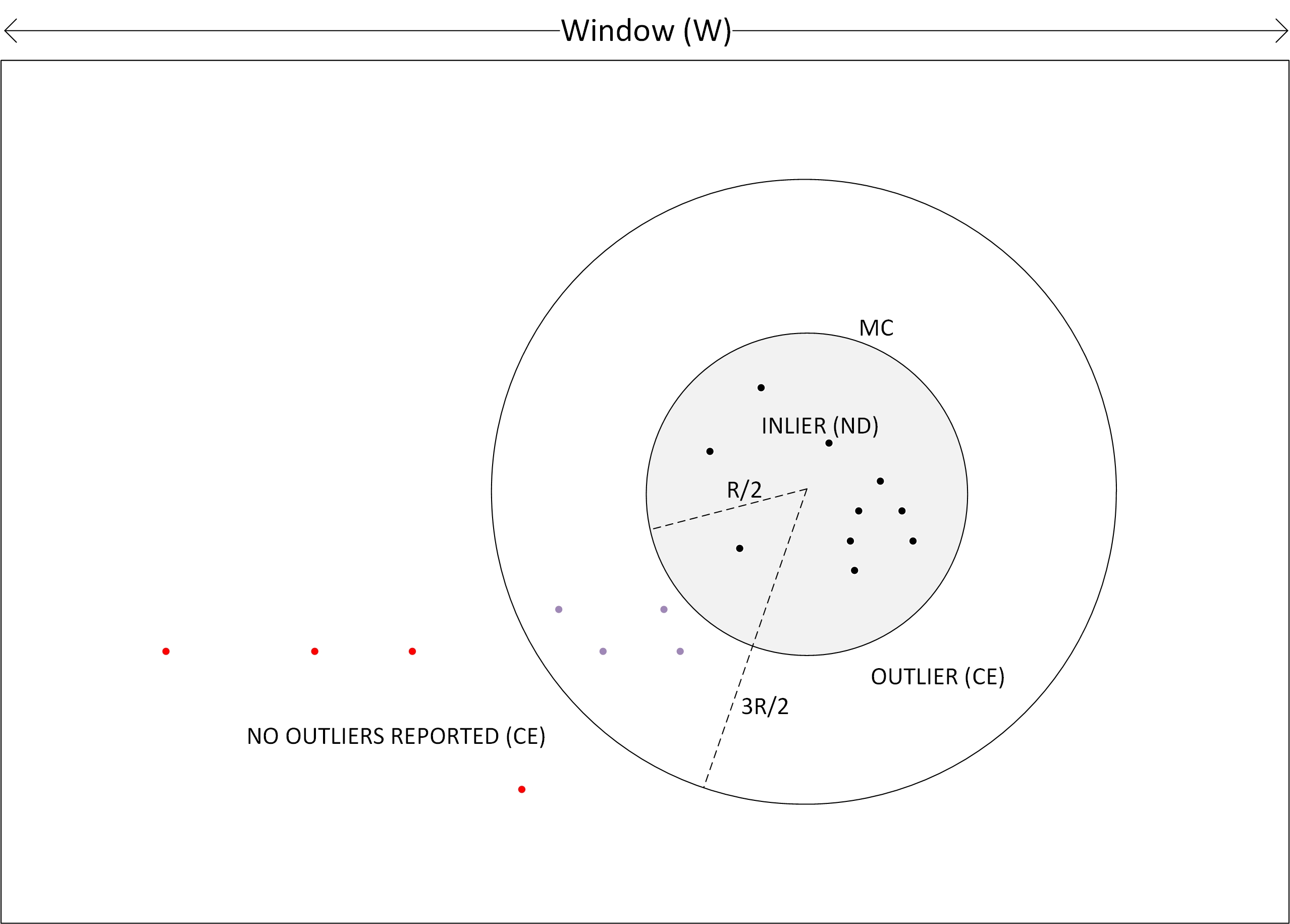}}
\caption{CE and ND Definition for \emph{DeepStreamCE}}
\label{fig4}
\end{figure}

The algorithm for the online phase is listed in Algorithm~\ref{algOn}. During the online phase, previously unseen instances arrive at the deep neural network. For each instance, the activations are extracted (line 2 of Algorithm~\ref{algOn}) and the deep neural network's prediction is stored (line 3 of Algorithm~\ref{algOn}). The activation layers are filtered and flattened to a 1D vector in the same way as during the offline phase (\textbf{flatten} -- line 6 of Algorithm~\ref{algOn}). The flattened activations are then processed through the autoencoder (\textbf{reduce} -- line 9 of Algorithm~\ref{algOn}) that was produced during the training phase. The reduced activation instance is added to the MCOD clusterer corresponding to the deep neural network’s predicted class for that instance (\textbf{addToClusterer} -- line 12 of Algorithm~\ref{algOn}) and the inlier/outlier information for that instance is obtained from the clusterer and a decision is made as to whether the instance is Non-Discrepancy (ND) or Concept Evolution (CE) (\textbf{analyse} -- line 13 of Algorithm~\ref{algOn}). During the \textbf{addToClusterer} phase, each unseen instance is applied to the MCOD clusterer associated with its class prediction. This clusterer only contains training data, so that no previously unseen instances affect the inlier/outlier decision. During the \textbf{analyse} phase, the inlier/outlier decision is obtained from the clusterer and transformed into a non-discrepancy or concept evolution decision. MCOD defines an inlier as $R/2$ of a cluster centre and an outlier as $3R/2$ as shown in Figure~\ref{fig4}.  For \emph{DeepStreamCE}, data points within $R/2$ (``INLIERS'') are reported as non-discrepancy (ND). Data points within the ranges of ``OUTLIER'' and ``NO OUTLIERS REPORTED'' are reported  as concept evolution (CE):  $p<R/2 \Rightarrow ND$ and $p > 3R/2 \Rightarrow CE$.

\begin{algorithm}
	\caption{Online Algorithm}
	\label{algOn}
\begin{algorithmic}[1]
        \Require {Pre-trained VGG16 DNN on 2 classes}
        \Require {Trained Autoencoder: ${A}$}
        \Require {Initialised MCOD clusterers: ${M}_{j}$}
        \Require {Specified activation layers: ${l}_{1}$...${l}_{n}$}
		\For {${u}$ = 1 \dots ${U}$}
		\State let $\mathbf{S}_{i}={v}_{l1}(u)_{i}...{v}_{lN}({u}_{i})$
		\State Get DNN prediction: ${j}$
		\EndFor
		\For {${l}$ = 1\dots${N}$}
		\State ${f}_{i}$ = \textbf{flatten}(${S}_{i}$)
		\EndFor
		\For {${u}$ = 1\dots${I}$}
		\State Reduce activations to 100: ${r}_{i}$ = \textbf{reduce}(${A}$,${f}_{i}$)
		\EndFor
		\For {${r}$ = 1\dots${I}$}
    		\State Add instance to stream: \textbf{addToClusterer}(${M}_{j}$,${r}_{ij}$)
		    \State Analyse for outlier: ${O}_{ij}$ = \textbf{analyse}(${M}_{j}$,${r}_{ij}$)
		\EndFor
		\State\Return ${O}_{ij}$
	\end{algorithmic} 
\end{algorithm}

\section{Experimental Study}
\label{sec:exp}
\subsection{Data Setup}
\label{sec:exp_datset}
The aim of \emph{DeepStreamCE} is to detect concept evolution. Therefore, we need to introduce a new class into the system, other than the classes it has been trained on. To achieve this, the deep neural network is only trained on two of the classes in the CIFAR-10 dataset, then a third class that the deep neural network has not been trained on from the CIFAR-10 dataset is introduced. The CIFAR-10 data set \cite{krizhevsky_learning_2009} consists of 10 different classes of $32 \times 32$ colour images. In total there are 50000 training images and 10000 test images. Table~\ref{tab3} provides a list of the classes, along with a higher granularity classification/type.    

\begin{table}[H]
\centering
\caption{CIFAR-10 Dataset Classes}
\begin{tabular}{ll}
\hline\noalign{\smallskip}
\label{tab3}
Class Name & Class Type\\
\hline\noalign{\smallskip}
airplane & Vehicle \\
automobile & Vehicle\\
bird & Animal \\
cat & Animal \\
deer & Animal\\
dog & Animal\\
frog & Animal \\
horse & Animal\\
ship & Vehicle \\
truck & Vehicle \\
\hline\noalign{\smallskip}
\end{tabular}
\end{table}

As listed in Table~\ref{tab3}, the CIFAR-10 data consists of four different types of vehicles and six different types of animals. The classes are mutually exclusive, however, some classes have more separation from each other. For instance, airplane and automobile are more similar to each other than the frog class.  This separation of types will be used to introduce concept evolution. The data setup specifications are defined as \emph{(NDclass,NDclass-CEclass)}.  Where the ND class are the non-discrepancy classes the neural network is trained on, and the CE class is the class that is introduced to simulate concept evolution.  The data setup specifications are split into two groups. The first group will utilise data consisting of two vehicle classes, then concept evolution will be introduced by applying unseen instances of a type of animal from the dataset.  The second group consists of class combinations that are perceived to have less separation between the classes. Airplane, ship and bird are selected for their similar backgrounds, giving overall image similarity. Ship, truck, automobile are selected as they are all transport. Cat, frog, deer are selected as they are all animals and cat, deer, horse are selected as they are four legged animals. The combinations of classes that the deep neural network will be trained on and the concept evolution classes are shown in Table~\ref{tab4}.

\begin{table}[H]
\centering
\caption{Concept Evolution Class Combinations}
\begin{tabular}{lll}
\hline\noalign{\smallskip}
\label{tab4}
\pbox{20cm}{Data\\Setup Name} & \pbox{20cm}{Trained Classes}& \pbox{20cm}{Concept Evolution\\Class}\\
\hline\noalign{\smallskip}
(airplane,automobile-frog)&airplane, automobile&frog\\
(ship,truck-cat)&ship, truck&cat\\
(airplane,truck-deer)&airplane, truck&deer\\
(ship,truck-bird)&ship, truck&bird\\
\hline\noalign{\smallskip}
(airplane,ship-bird)&airplane, ship&bird\\
(ship,truck-automobile)&ship, truck&automobile\\
(cat,frog-deer)&cat, frog&deer\\
(cat,deer-horse)&cat, deer&horse\\
\hline\noalign{\smallskip}
\end{tabular}
\end{table}

\subsection{Deep Neural Network}
\label{sec:exp_dnn}
\begin{figure}[H]
\centering
\includegraphics[width=8cm]{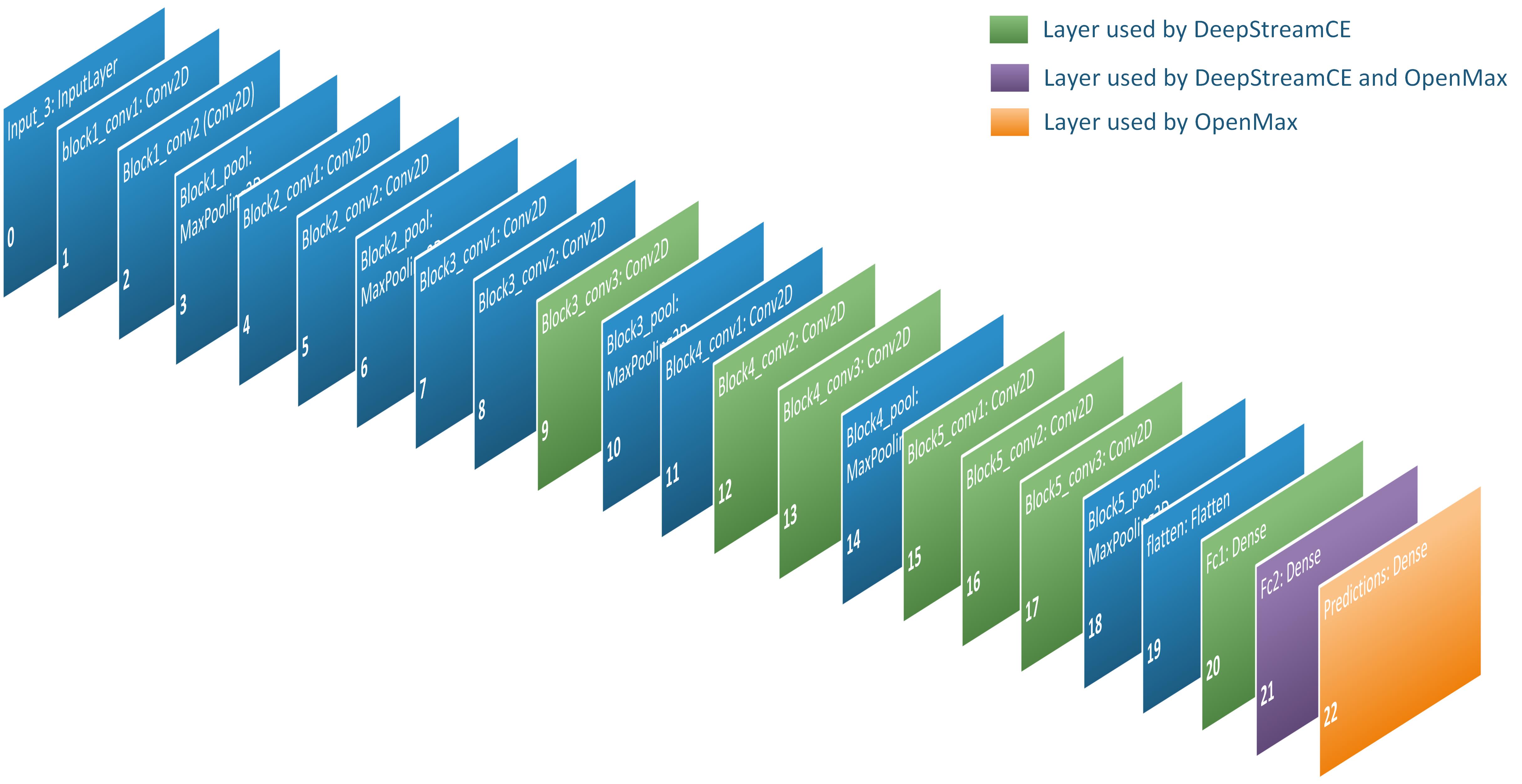}
\caption{VGG16 Network Representation with layer names as reported from Keras}
\label{fig5}
\end{figure}

The deep neural network that the system operates on is a widely used base model, VGG16 \cite{simonyan_very_2015}. This model was originally designed and trained on imagenet; we have trained it on 2 classes at a time from the CIFAR-10 dataset. The system is designed such that it is not limited to only being able to use VGG16 as the deep neural network, however, for our experiments we have selected VGG16 as it is a smaller efficient network and due to the amount of activations that are produced, this is a manageable size for our initial experiments. Figure~\ref{fig5} shows the layers in the VGG16 network, which are numbered from the top starting with 0. The following 8 layers are being used: 9,12,13,15,16,17,20,and 21. Using a small network allows a good representative number of layers to be selected to facilitate future work in analysing the usefulness of the layers. These particular layers were selected as the closer the hidden layer is to the end of the network, the more feature information it contains. Therefore, the final convolutional layer, prior to each pooling layer was selected to provide maximum information rather than the pooling layers. The open classification technique we are comparing \emph{DeepStreamCE} to is OpenMax as described in Section~\ref{sec:relwk_omax}, OpenMax uses the final two layers of the network -- the fc2 (Dense) layer and the predictions (Dense) layer.

\subsection{Experimental Setup}
\label{sec:exp_expset}
For the experimental setup, four class combinations are being used as defined in Table~\ref{tab3}.  For each of these data setup specifications, a parameter investigation is conducted to investigate the effect that the MCOD parameters $k$ (the minimum number of neighbours required to form an MCOD micro cluster) and $R$ (the radius of the micro cluster) has on the Recall and Precision.  From this, $k$ and $R$ parameters are deemed to be the most influential on \emph{DeepStreamCE}. The \emph{DeepStreamCE} experimental setup is trained on 5000 instances per class and utilises 500 unseen instances per run, 250 of which are non-discrepancy instances and 250 of which are concept evolution instances. The unseen instances are selected randomly from the test data of 1000 instances. Any test instances that are wrongly classified by the network are removed for experimental purposes. These runs are repeated 4 times and averages are taken. For OpenMax, all the non-discrepancy and concept evolution instances for that data setup are applied to the system in one run. (2000 non-discrepancy instances and 1000 concept-evolution instances).  Similarly, any test instances that are wrongly classified by the network (softmax layer) are removed.

Each data setup specification is applied to the OpenMax open-classification method to detect unknown classes\cite{bendale_towards_2016}. The OpenMax experimental setup has been modified to work with the VGG16 deep neural network and our data setup specifications as defined in Table~\ref{tab5}. This was modified with the assistance of source code from the original paper \cite{bendale_abhijitbendaleosdn_2020} and a wrapper implementation from \cite{neupane_aadeshnpnosdn_2019}.

\begin{table}[!htp]
\centering
\caption{Experimental Setup}
\label{tab5}
\begin{tabular}{llll}
\hline\noalign{\smallskip}
Data Setup & Layer Filtering & \pbox{20cm}{Activation\\Reduction}  & \pbox{20cm}{Activation\\Analysis}\\
\hline\noalign{\smallskip}
(airplane,automobile-frog)&8 layers&autoencoder&MCOD\\
(ship,truck-cat)&8 layers&autoencoder&MCOD\\
(airplane,truck-deer)&8 layers&autoencoder&MCOD\\
(ship,truck-bird)&8 layers&autoencoder&MCOD\\
(airplane,ship-bird)&8 layers&autoencoder&MCOD\\
(ship,truck-automobile)&8 layers&autoencoder&MCOD\\
(cat,frog-deer)&8 layers&autoencoder&MCOD\\
(cat,deer-horse)&8 layers&autoencoder&MCOD\\
(airplane,automobile-frog)&penultimate layer&N/A&OpenMax\\
(ship,truck-cat)&penultimate layer&N/A&OpenMax\\
(airplane,truck-deer)&penultimate layer&N/A&OpenMax\\
(ship,truck-bird)&penultimate layer&N/A&OpenMax\\
(airplane,ship-bird)&penultimate layer&N/A&OpenMax\\
(ship,truck-automobile)&penultimate layer&N/A&OpenMax\\
(cat,frog-deer)&penultimate layer&N/A&OpenMax\\
(cat,deer-horse)&penultimate layer&N/A&OpenMax\\
\hline\noalign{\smallskip}
\end{tabular}
\end{table}

\subsection{Evaluation Metrics}
\label{sec:exp_evalmet}
The following metrics are computed: true positives (TP), false positives (FP), true negatives (TN) and false negatives (FN).  True positives are defined as images that belong to the new class and were correctly identified as concept evolution. False positives are defined as images that belong to an existing class and were incorrectly identified as concept evolution. True negatives are defined as images that belong to an existing class and were correctly identified as such. False negatives are defined as images that belong to a new class and were incorrectly identified as belonging to an existing class. From this data, we calculate Precision, Recall and F-Measure as defined in Table~\ref{tab6}.
The OpenMax paper \cite{bendale_towards_2016} uses F-measure to evaluate open-set performance as it is better than using accuracy, because it is not inflated by true negatives. It combines precision and recall -- it is the harmonic mean. \cite{scheirer_toward_2013} For a given threshold on OpenMax/SoftMax probability values, they compute true positives, false positives and false negatives over the entire dataset.  To compare \emph{DeepStreamCE} with OpenMax, we are using a modified version of OpenMax that we have adapted to work with our data and deep neural network. The parameters of alpha and tail required modification for our data and optimum values for these were found empirically to be 2 and 9, respectively.  See the code availability section for a link to the source code for the implementation of this research.

\begin{table}[H]
\caption{Performance Measures}
\label{tab6}
\begin{tabular}{lll}
\hline\noalign{\smallskip}
Name & Description & Formula  \\
\noalign{\smallskip}\hline\noalign{\smallskip}
Precision & \pbox{20cm}{Ratio of CE instances that are\\declared as outliers amongst all outliers} &$\frac{TP}{TP+FP}$\\
& & \\
Recall & \pbox{20cm}{Ratio of CE instances that are\\declared as CE amongst all CE instances} &$\frac{TP}{TP+FN}$ \\
& & \\
F-Measure & \pbox{20cm}{The harmonic mean\\(combination of Precision and Recall)} & $\textstyle\frac{2 \times Precision \times Recall}{Precision + Recall}$ \\
\noalign{\smallskip}\hline
\end{tabular}
\end{table}

\section{Experimental Results}
\label{sec:res}
\subsection{Parameter Investigation Results}
\label{sec:res_paramRes}
\begin{figure*}[htbp]
\begin{subfigure}{.5\textwidth}
\centerline{\includegraphics[width=\textwidth]{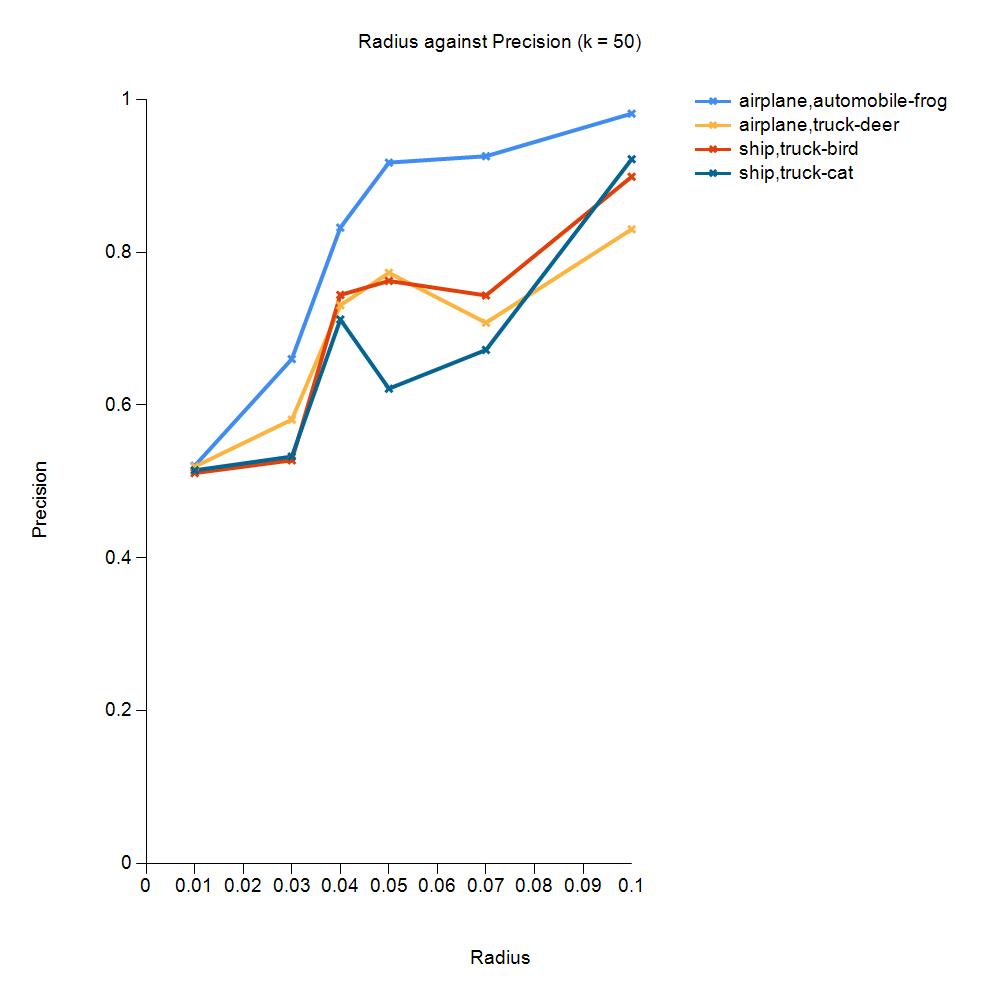}}
\caption{Variation of Precision with $R$}
\label{fig:rad_sub1}
\end{subfigure}
\begin{subfigure}{.5\textwidth}
\centerline{\includegraphics[width=\textwidth]{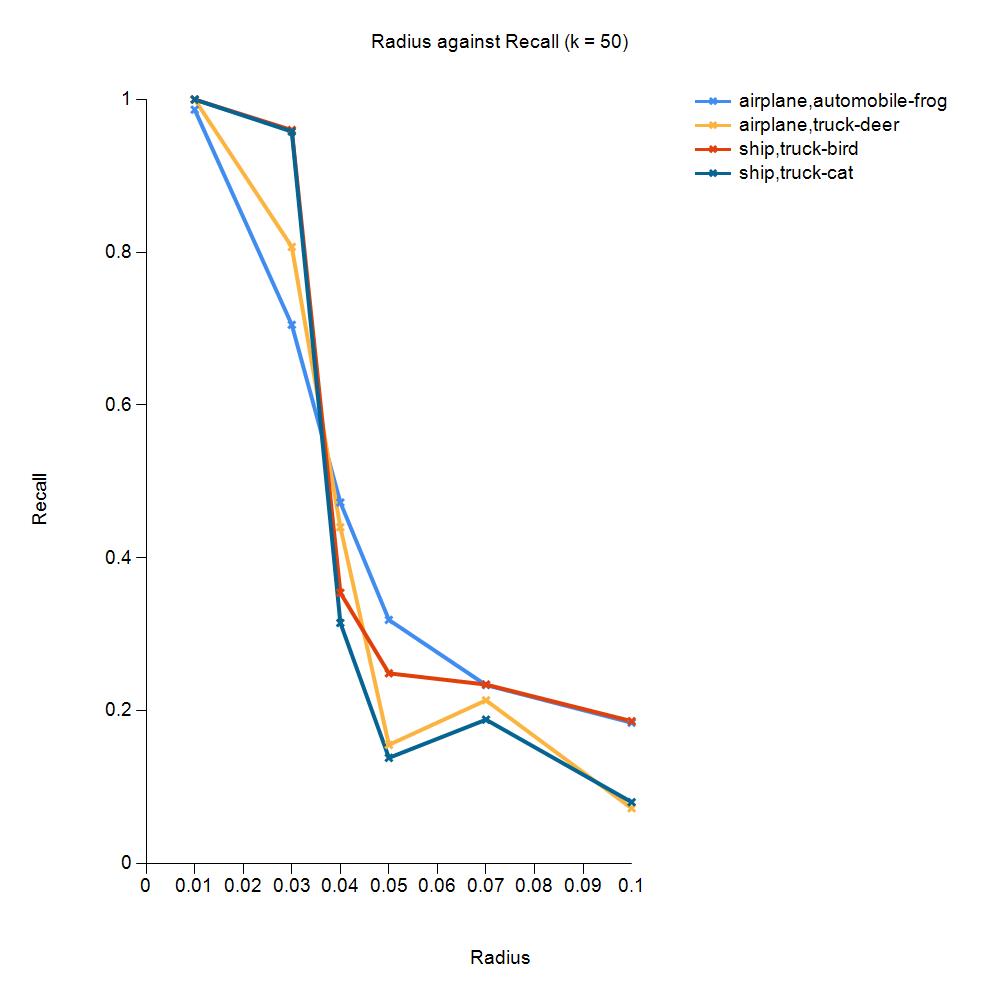}}
\caption{Variation of Recall with $R$}
\label{fig:rad_sub2}
\end{subfigure}
\caption{Variation of Precision and Recall with ${R}$ (${k}$ = 50)}
\label{fig:rad}
\end{figure*}

\begin{figure*}[htbp]
\begin{subfigure}{.5\textwidth}
\centerline{\includegraphics[width=\textwidth]{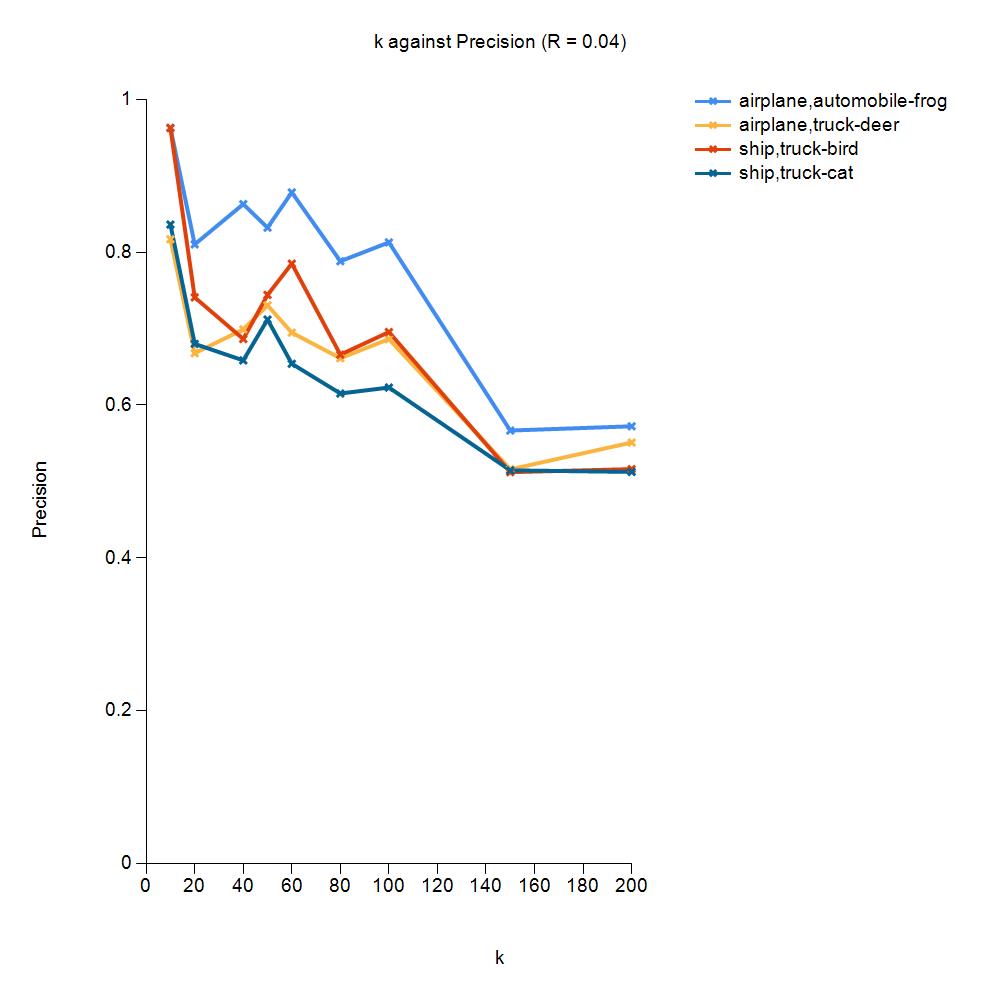}}
\caption{Variation of Precision with $k$}
\label{fig:k_sub1}
\end{subfigure}
\begin{subfigure}{.5\textwidth}
\centerline{\includegraphics[width=\textwidth]{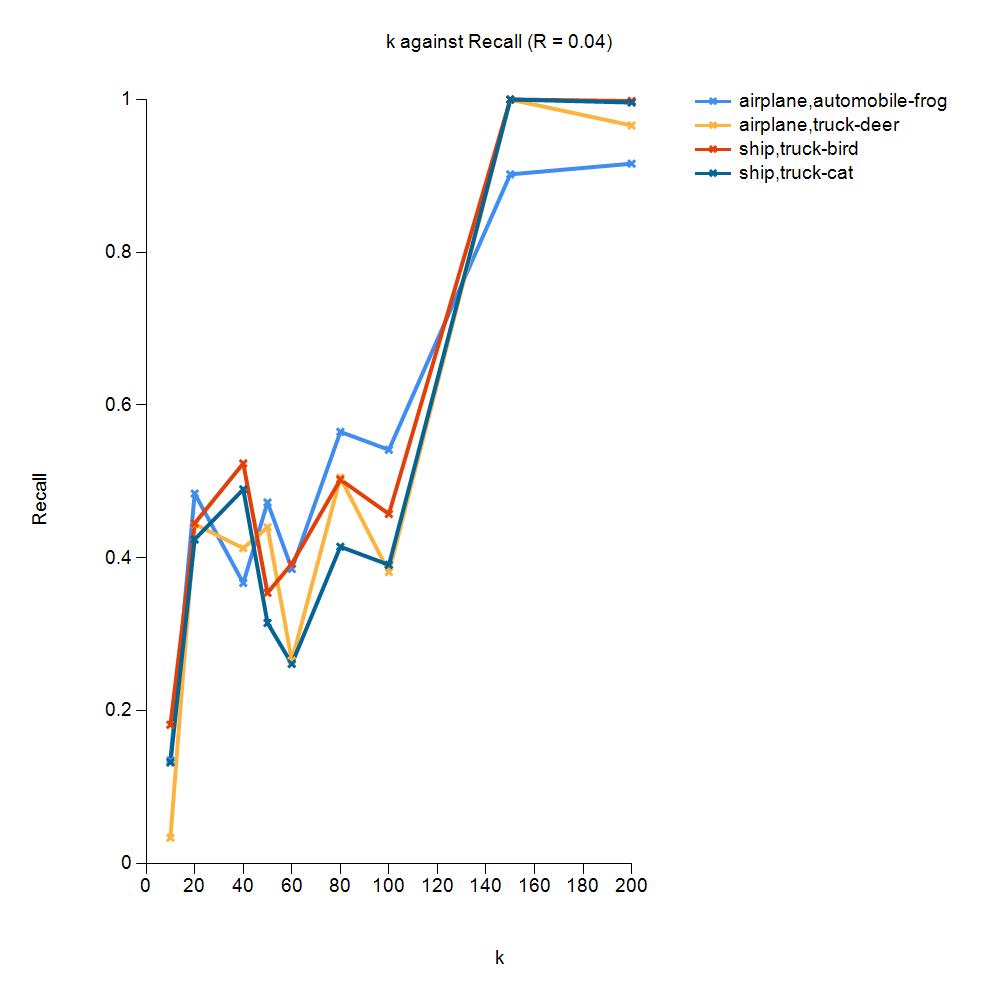}}
\caption{Variation of Recall with $k$}
\label{fig:k_sub2}
\end{subfigure}
\caption{Variation of Precision and Recall with ${k}$ (${R}$ = 0.04)}
\label{fig:k}
\end{figure*}

The parameters $k$ (the minimum number of neighbours required to form an MCOD micro cluster) and $R$ (the radius of the micro cluster) were varied between 10 and 200 and 0.01 and 0.1, respectively.  Figures~\ref{fig:rad} and~\ref{fig:k} show the effect these changing parameters has on Precision and Recall. Figure~\ref{fig:rad} demonstrates that the smaller the radius, the better the recall. However, if the radius is set too small the precision will drop, suggesting that 0.04 would produce a balance of Precision and Recall.  Figure~\ref{fig:k} demonstrates that the higher the value of $k$, the better the recall without a large drop in Precision, suggesting that $k$ could be set to 80 with only a small drop in precision.  

\newpage

\subsection{DeepStreamCE and OpenMax Results}
\label{sec:res_dsRes}
For comparison of \emph{DeepStreamCE} to OpenMax,  the MCOD parameters of $k$ = 80 and $R$ = 0.04 have been used as suggested by the experimentation as reported in Section~\ref{sec:res_paramRes}.  \emph{DeepStreamCE} is compared with OpenMax via Precision, Recall and F-Measure.  Table~\ref{tab:res} shows the results for these measures.

\begin{table}[H]
\centering
\caption{Experimental Results for \emph{DeepStreamCE} (DS) and OpenMax (OM)}
\label{tab:res} 
\begin{tabular}{lllllll}
\hline\noalign{\smallskip}
\pbox{20cm}{Data\\Setup} & \pbox{20cm}{DS\\Precision} & \pbox{20cm}{DS\\Recall} & \pbox{20cm}{DS\\F-Meas} & \pbox{20cm}{OM\\Precision} & \pbox{20cm}{OM\\Recall} & \pbox{20cm}{OM\\F-Meas}\\
\noalign{\smallskip}\hline\noalign{\smallskip}
(airplane,automobile-frog)&0.788&0.565&\textbf{0.638}&0.700&0.178&0.284\\
(ship,truck-cat)&0.615&0.414&\textbf{0.436}&0.750&0.096&0.170\\
(airplane,truck-deer)&0.661&0.505&\textbf{0.518}&0.085&0.079&0.081\\
(ship,truck-bird)&0.666&0.502&\textbf{0.531}&0.735&0.089&0.158\\
\noalign{\smallskip}\hline\noalign{\smallskip}
(airplane,ship-bird)&0.527&0.192&\textbf{0.281}&0.191&0.225&0.206\\
(ship,truck-automobile)&0.361&0.050&0.088&0.640&0.057&\textbf{0.104}\\
(cat,frog-deer)&0.594&0.848&\textbf{0.698}&0.000&0.000&0.000\\
(cat,deer-horse)&0.527&0.271&\textbf{0.357}&0.317&0.128&0.182\\
\noalign{\smallskip}\hline
\end{tabular}
\end{table}

For the first group of data, where the class combinations consist of vehicles as the trained classes and an animal as the concept evolution class, the results for \emph{DeepStreamCE} indicate that the data setup of (airplane,automobile-frog) obtained the best precision and recall, with (ship,truck-cat) obtaining the lowest results. This suggests that the frog is more distinguishable from transport than cat. 
For OpenMax, the results show that the F-Measure is lower on all data setup specifications than \emph{DeepStreamCE}, because of low Recall rates. This means that a high percentage of unknown classes are assigned as known classes. The OpenMax F-Measure scores follow the trend of the \emph{DeepStreamCE} F-Measure scores in that the most separated data setup of (airplane,automobile-frog) also displays the highest F-Measure score.  Data setup (airplane,truck-deer) shows very low Precision and Recall, this was also one of the lower scoring data setup specifications on \emph{DeepStreamCE}.
In the second group of data setup specifications, for \emph{DeepStreamCE}, the F-Measure results were generally lower than the first group as expected due to the more similar nature of the data.  However, (cat,frog-deer) is the highest of all data setup specifications with an F-Measure of 0.698 and a high recall of 0.848 reported.  OpenMax reports higher F-Measure results for (ship,truck-automobile) all transport categories, although this has a very low recall -- so it identified automobile as the unknown class, but also identified automobile as known classes a high number of times. Both \emph{DeepStreamCE} and OpenMax struggled to identify concept evolution when all classes were vehicles. OpenMax scored zero on (cat,frog-deer) where it did not identify any instances as unknown -- it could not identify a difference between any of these classes. From these results, it can be seen that \emph{DeepStreamCE} outperforms OpenMax in the scenario of detecting concept evolution for the data setup specifications provided.
Using the Wilcoxon Signed-Rank test, the difference between the F-Measure of \emph{DeepStreamCE} and that of OpenMax over the eight tested cases is statistically significant. The $p$-value = 0.01563 which is less than 0.05 significance level, suggesting the acceptance of the alternative hypothesis that true location shift is not equal to 0. 

\subsection{Detection Time Analysis}
\label{sec:res_dsTime}
In a streaming scenario, the amount of time that is taken to process one instance is of interest. Table~\ref{tab10} below shows the average, minimum and maximum time taken to process instances for \emph{DeepStreamCE} and OpenMax. The average speed of detection for \emph{DeepStreamCE} is 324ms per instance. This is the time per instance, measured in batches of 100 instances. The system is running on 64vCPUs, 416GB RAM, 4 x NVIDIA Tesla T4 GPUs. This is the duration it takes to select the layers, flatten the activations, reduce them in the autoencoder and process them through the MCOD clustering algorithm to produce an outlier result. The variation between the time taken on all runs is within 55ms and the run times stay consistent when varying ${R}$ and ${k}$. The average time taken for OpenMax to calculate its outcome for each instance is 257ms. This is the time it takes to compute the mean activation vector for an instance, apply this to the Weibull distribution of the mean activation vectors for each class and re-calibrate the output decision to allow an `unknown' classification. The results show that OpenMax is faster than \emph{DeepStreamCE}. However, OpenMax considers much less activation data than \emph{DeepStreamCE} and performs less computations as it based on probability rather than activation reduction via an autoencoder and cluster processing. OpenMax performs considerably lower in the concept evolution detection performance than \emph{DeepStreamCE} and only provides a small decrease in execution time.

\begin{table}[H]
\caption{Time taken to process an instance in ms}
\centering
\label{tab10} 
\begin{tabular}{llllllllll}
\hline\noalign{\smallskip}
System & Average & Min & Max\\
\noalign{\smallskip}\hline\noalign{\smallskip}
DeepStreamCE&324&306&361\\
OpenMax&\textbf{257}&256&260\\
\noalign{\smallskip}\hline
\end{tabular}
\end{table}

\section{Conclusion and Future Work}
\label{sec:concl}
The experiments have shown that detection of concept evolution utilising deep neural network activations via streaming detection methods is a viable approach. This was proven using two separated types of classes (transport and animals) from the CIFAR-10 dataset. The effectiveness of this was compared to OpenMax where \emph{DeepStreamCE} outperformed OpenMax. The value of the radius, ${R}$ and the number of neighbours in a cluster (${k}$) of the MCOD clusterer are significant factors with regards to the concept evolution decision, with the potential to increase the Recall with only a small decrease in Precision. This research has demonstrated an introduction into utilising deep neural network activations in a streaming environment to detect concept evolution. Further directions of study are: (1) expanding the data into using more classes and less separated classes, (2) extending the analysis into concept drift and adversarial detection, using larger neural network models, (3) investigating which are the optimum network layers to use and experimenting on data other than images and different types of deep neural networks, as the system becomes data agnostic once the activations are utilised instead of the input data.

\begin{acknowledgements}
The authors thank the Google Cloud Platform Research Credits scheme for enabling this research via the use of cloud resources and software utilised from MOA \cite{bifet_moa_2010} and Py4J \cite{dagenais_py4j_nodate}. 
\end{acknowledgements}

\bibliographystyle{spmpsci}      
\bibliography{references}   

%
%

\end{document}